\documentclass[runningheads]{llncs}
\usepackage{url}
\usepackage{xcolor}
\definecolor{c1}{rgb}{0,0,1} 
\definecolor{c2}{rgb}{0.1,0.1,0.1} 
\definecolor{c3}{rgb}{0.2,0.2,0.7} 
\usepackage[colorlinks=true,citecolor=c3]{hyperref}

\usepackage{wrapfig}

\usepackage{microtype}
\usepackage{graphicx}
\usepackage{subfigure}
\usepackage{booktabs} 

\usepackage[utf8]{inputenc} 
\usepackage[T1]{fontenc}    
\usepackage{url}            
\usepackage{booktabs}       
\usepackage{amsfonts}       
\usepackage{nicefrac}       
\usepackage{microtype}      

\usepackage{subfigure}
\usepackage{booktabs} 

\usepackage{amsmath,amssymb,bm}

\usepackage{shortbold}

\usepackage{algorithm}
\usepackage{algorithmic}
\usepackage[misc]{ifsym}

\begin{document}
\title{Neural Control Variates for Monte Carlo Variance Reduction}
%
%
\toctitle{Neural Control Variates for Monte Carlo Variance Reduction}

\author{Ruosi Wan\inst{1} \and
Mingjun Zhong\inst{2} \and
Haoyi Xiong\inst{3} \and
Zhanxing Zhu\inst{1,4,5}\Letter}
\authorrunning{R. Wan et al.}
%
\tocauthor{Ruosi~Wan(Peking University),
Mingjun~Zhong(University of Lincoln),
Haoyi~Xiong(Baidu Inc.),
Zhanxing~Zhu(Peking University)}

\institute{ 
Center for Data Science, Peking University, Beijing, China \email{\{ruoswan,zhanxing.zhu\}@pku.edu.cn}\and
School of Computer Science, University of Lincoln, United Kingdom \email{mzhong@lincoln.ac.uk}\and 
Big Data Lab, Baidu Inc., Beijing, China \email{xionghaoyi@baidu.com}\and
School of Mathematical Sciences, Peking University, Beijing, China \and
Beijing Institute of Big Data Research, Beijing, China
}

\maketitle              
\begin{abstract}
In statistics and machine learning, approximation of an intractable integration is often achieved by using the unbiased Monte Carlo estimator, but the variances of the estimation are generally high in many applications. Control variates approaches are well-known to reduce the variance of the estimation. These control variates are typically constructed by employing predefined parametric functions or polynomials, determined by using those samples drawn from the relevant distributions. Instead, we propose to construct those control variates by learning neural networks to handle the cases when test functions are complex. In many applications, obtaining a large number of samples for Monte Carlo estimation is expensive, the adoption of the original loss function may result in severe overfitting when training a neural network. This issue was not reported in those literature on control variates with neural networks. We thus further introduce a constrained control variates with neural networks to alleviate the overfitting issue. We apply the proposed control variates to both toy and real data problems, including a synthetic data problem, Bayesian model evidence evaluation and  Bayesian neural networks. Experimental results demonstrate that our method can achieve significant variance reduction compared to other methods.

\keywords{Control variates  \and Neural networks \and Variance reduction  \and Monte Carlo method.}
\end{abstract}
\section{Introduction}
Most of modern machine learning and statistical approaches focus on modelling complex data, where manipulating high-dimensional and multi-modal probability distributions is of great importance for model inference and learning. Under this circumstance, evaluating the expectation of certain function $f(\thetaB)$ with respect to a probability distribution $p(\thetaB)$ is ubiquitous, 
\begin{equation}
\mu = \Ebb_{\thetaB \sim p(\thetaB)} [f(\thetaB)] = \int f(\thetaB) p(\thetaB) d \thetaB, 
\end{equation}
where the random variable of interest $\thetaB \in \Rbb^D$ is typically high-dimensional. 

However, in complex models, the integration is often analytically intractable. This drives the development of sophisticated Monte Carlo methods to facilitate efficient computation~\cite{robert2004monte}. The Monte Carlo method is naturally employed to approximate the expectation, i.e.,
\begin{equation}
\mu  \approx  \frac{1}{n} \sum_{i=1}^n f(\thetaB_i),
\end{equation}
where $\{\thetaB_i \}_{i=1}^n$ are samples drawn from the distribution $p(\thetaB)$.  According to the central limit theorem, this estimator converges to $\mu$ at the rate $O(1/\sqrt{n})$. For high-dimensional and complex models,  \emph{when $p(\thetaB)$ is difficult to sample from}~\cite{neal2012bayesian}  or \emph{the test function $f$ is expensive to evaluate}~\cite{higdon2015bayesian},  a ``large-$n$'' estimation is  computationally prohibited.  This directly leads to a high-variance estimator. Therefore, with a limited number of samples,  how to reduce the variance of Monte Carlo estimations emerges as an essential issue for its practical use. 

Along this line, various variance reduction methods have been introduced in the literature of statistics and numerical analysis. One category aims to develop appropriate samplers for variance reduction, including importance sampling and its variants~\cite{cornuet2012adaptive}, stratified sampling techniques~\cite{rubinstein2016simulation}, multi-level Monte Carlo~\cite{giles2013multilevel} and other sophisticated methods based on Markov chain Monte Carlo (MCMC)~\cite{robert2004monte}. Another category of variance reduction methods is called control variates~\cite{assaraf1999zero,mira2013zero,oates2016controlled,oates2017control,liu2017action,tucker2017rebar}. These methods take advantage of random variables with known expectation values, which are negatively correlated with the test function under consideration. Control variates techniques  can fully employ the available samples to reduce the variance, which is popular due to its efficiency and effectiveness. 

However, existing control variates approaches have several limitations. Firstly, most existing methods use a linear or quadratic form to represent the control function~\cite{mira2013zero,oates2016controlled}. Although these control functions have closed forms, the representation power of them is very limited particularly when the test function $f(\thetaB)$ is complex and non-linear. Control functionals were proposed recently to tackle this problem~\cite{oates2017control}. However, these estimators may significantly suffer from a curse of dimensionality~\cite{oates2016convergence}. Secondly, when the available samples are scarce, optimizing the control variates only based on a small, number of samples might overfit, which means that it is difficult to generalize on the samples obtained later. These restrictions limit their practical performance. 

In order to overcome the first issue, some works~\cite{liu2017action, tucker2017rebar} employed neural networks to represent the control variates, utilizing the capability of a neural network to represent a complex test function. We name these methods as ``Neural Control Variates'' (\textbf{NCV}). Unfortunately, in the scenario of learning neural networks,  applying the commonly used loss function to reduce variance causes severe overfitting issue, particularly when available training sample size is small. Therefore, we introduce ``Constrained Neural Control Variates'' (\textbf{CNCV}) which makes constraints on the control variates for alleviating the over-fitting issue.   
Our method is particularly suitable for the cases when the sample space is high-dimensional or the samples from $p(\thetaB)$ is hard to obtain.  We demonstrate the effectiveness of our approach on both synthetic and real machine learning tasks, including 1) expectation of a complex function under the mixture of Gaussian distributions,  2) Bayesian model evidence evaluation and 3) Bayesian neural networks.  We show that CNCV achieved the best performance comparing to the state-of-the-art methods in literature.  



\section{Control Variates}
\label{sec:cv}
The generic control variates aims to estimate the expectation $\mu = \Ebb_{p(\thetaB)} [f(\thetaB)]$ with reduced variance. The principle behind the control variates relies on constructing an auxiliary function $\tilde{f}(\thetaB) = f(\thetaB) + g(\thetaB)$ such that
\begin{equation}
\Ebb_{p(\thetaB)} [g(\thetaB)] = 0.
\end{equation}
Thus the desired expectation can be replaced by that of the auxiliary function
\begin{equation}
\mu = \Ebb_{p(\thetaB)} [f(\thetaB)] = \Ebb_{p(\thetaB)} [\tilde{f}(\thetaB)].
\end{equation}
It is possible to obtain a variance-reduced Monte Carlo estimator by \emph{selecting or optimizing} $g(\thetaB)$ 
so that the variance $\Vbb_{p(\thetaB)} [\tilde{f}(\thetaB)] < \Vbb_{p(\thetaB)} [f(\thetaB)]$. Intuitively,  variance reduction can be achieved 
when $g(\thetaB)$ is negatively correlated with $f(\thetaB)$ under $p(\thetaB)$, since much of the randomness “cancels out” in the auxiliary function $\tilde{f}(\thetaB)$. 

The selection of an appropriate form of control function $g(\thetaB)$ is crucial for the performance of variance reduction.  A tractable class of so called zero-variance control variates was proposed in~\cite{assaraf1999zero,mira2013zero}. Those control variates are expressed as a function of the gradient of the log-density, $\nabla_{\thetaB} \log p(\thetaB)$, i.e. the score function $\sB(\thetaB)$. Concretely, it has the following form
\begin{equation}
g(\bm{\theta}) = \Delta_{\thetaB} Q(\bm{\theta}) + \nabla_{\thetaB} Q(\bm{\theta}) \cdot \nabla_{{\thetaB}} \log (p(\bm{\theta})),
\label{eq:laplacecv}
\end{equation}
where the gradient operator $\nabla_{\thetaB} = [\partial/\partial \theta_1, \dots, \partial/\partial \theta_D]^T$, the Laplace operator $\up_{\thetaB} = \sum_{i=1}^D \partial^2 / \partial \theta_i^2$, and ``$\cdot$'' denotes the inner product. The function $Q(\thetaB)$ is often referred to as the trial function. The target is now to find a trial function so that $g(\bm{\theta})$ and $f(\bm{\theta})$ are negatively correlated. This could thus reduce the variance of the Monte Carlo estimation.  
As the trial function could be arbitrary under those mild conditions given in ~\cite{mira2013zero}, a parametric function could be used for $Q(\thetaB)$. For example, when $Q(\thetaB)  = \aB^T \thetaB$, which is a first degree polynomial function, the auxiliary function becomes 
\begin{equation}
\tilde{f}(\thetaB) = f(\thetaB) + \aB^T \sB(\thetaB)
\end{equation}
 as was proposed in ~\cite{mira2013zero}. The optimal choice of the parameter $\aB$ that minimizes the variance of $\tilde{f}(\thetaB)$ is 
$
\aB = - \SigmaB_{\sB \sB}^{-1} \sigmaB(\sB, f), \label{eq:a} 
$
where $\SigmaB_{\sB \sB} = \Ebb [\sB \sB^T]$,  $\sigmaB(\sB, f) = \Ebb[\sB f]$. Obviously, the representation power of these polynomials is limited, and therefore control functionals have been proposed recently where the trial function is stochastic. For example, the trial function could be a kernel function~\cite{oates2017control}.
In order for using these control variates, we firstly estimate these required parameters in the trial function by using some training samples $\{ \thetaB_i \}_{i=1}^n$. Then the learned control variates can be used for test samples.

However, there are several drawbacks of the current zero-variance techniques:
\begin{itemize}
 \item \emph{Dilemma between effectiveness and efficiency.} Although increasing the order of polynomial could potentially increase the representation power and reduce more variance, the number the parameters needs to be learned would grow exponentially. As pointed out by~\cite{mira2013zero}, when quadratic polynomials are used, $Q(\thetaB) = \aB^T \thetaB +\thetaB^T \BB \xB /2$, the number of parameters will be $D(D+3)/2$. Thus, finding the optimal coefficients requires dealing with $\SigmaB_{\sB \sB}$ which is a matrix
of dimension of order $D^2$. Similar issue occurs when employing the control functionals. This makes the use of high order polynomials computationally expensive when faced with high-dimensional sampling spaces. 
\item \emph{Poor generalization with small sample size.} With small sizes of training samples and complex $p(\thetaB)$,  the learned control variates could potentially overfit the training samples, i.e. generalize poorly over new samples.  This is because a small size of training samples might be insufficient for representing the full distributional information of $p(\thetaB)$.  
\end{itemize}

These limitations motivate the development of neural control variates and a novel loss function to alleviate overfitting issue when learning the neural control variates, which will be elaborated below.

\section{Neural Control Variates}
Firstly, we focus on alleviating  the dilemma between effectiveness and efficiency on designing control variates in  high-dimensional sample space. 
To this end, the trail function is designed as a neural network~\cite{liu2017action,tucker2017rebar}, we name this strategy as neural control variates (NCV). 
Equipped with neural network, their excellent capability of representing complex functions and overcoming the curse of dimensionality can be fully employed in high-dimensional scenarios~\cite{lecun2015deep}. 

Instead of relying on the control variates \eqref{eq:laplacecv} introduced in~\cite{mira2013zero}, we use the following Stein control variates based on Stein identity~\cite{stein1972bound,oates2017control}, 
\begin{equation}\label{eq:steincv}
g(\bm{\theta}) = \nabla_{\thetaB} \cdot \Phi(\bm{\theta}) + \Phi(\bm{\theta}) \cdot \nabla_{\thetaB} \log (p(\bm{\theta})),
\end{equation}
where $\Phi(\thetaB)$ is the trial function. Note that in order for $\Ebb[g(\thetaB)] =0$, we assume mild zero boundary conditions on $\Phi$, such that $p(\theta)\Phi(\theta)=0$ at the boundary or $\lim_{\|x\|\rightarrow\infty}p(\theta)\Phi(\theta)=0$ ~\cite{mira2013zero,liu2016stein01,oates2017control}. Compared with Eq~\eqref{eq:laplacecv}, Stein control variates is preferred due to its computational advantages since evaluating the second order derivatives of the trial function is avoided.  Note that when the trial function $Q(\thetaB)$ is a linear or quadratic polynomial, the Stein trial function $\Phi(\thetaB)$ is constant or linear, respectively.  

We now represent the trial function $\Phi(\thetaB)$ by a neural network $\Phi(\thetaB; \wB)$ parameterized by the weights $\wB$. The control function becomes $g(\thetaB; \wB) = \nabla_{\thetaB} \cdot \Phi(\bm{\theta}; \wB) + \Phi(\bm{\theta}; \wB) \cdot \nabla_{\thetaB} \log p(\bm{\theta})$. In order for variance reduction, we solve the following optimization problem 
\begin{equation}
\min_{\wB} \Vbb_{p(\thetaB)} [f(\thetaB) + g(\thetaB; \wB)],
\label{eq:oraclevar}
\end{equation}
which does not have a closed-form in general. Typically, it is assumed that the variance could be approximated by using independent Monte Carlo samples and so the optimization problem is given by
\begin{align}
\label{eq:var}
\min_{\wB} \frac{1}{n}\sum_{i=1}^n [f(\thetaB_i) + g(\thetaB_i; \wB)]^2  -  (\mu_0+\mu_{g})^2,
\end{align} 
where $\mu_0=E(f(\theta))$ and $\mu_g=E(g(\thetaB; \wB))=0$.
Instead, the following optimization problem will be solved 
\begin{align}
\label{eq:empvar}
\min_{\wB} \frac{1}{n}\sum_{i=1}^n [f(\thetaB_i) + g(\thetaB_i; \wB)]^2  
\end{align}
where $\{ \thetaB_i \}_{i=1}^n$ are samples drawn from $p(\thetaB)$.
Standard back-propagation techniques and stochastic gradient descent (SGD) can then be adopted to obtain the optimal weights of the neural networks. 

Unfortunately,  when the distribution $p(\thetaB)$ is high-dimensional and multi-modal, such as Bayesian neural networks, it would be very expensive to draw many samples for training control variates. Given a limited computational budget, it typically produces a rather small number of samples that are not sufficient for learning the control variates. Consequently, the learned parameters for control variates can easily overfit over the training samples, and thus could not generalize well on new samples drawn from $p(\thetaB)$. This overfitting phenomenon was not noticed  in ~\cite{liu2017action,tucker2017rebar}, since the considered applications in their scenarios only involve either a simple probability distribution $p(\thetaB)$ or simple target function $f(\thetaB)$. 

Therefore, in the following, we propose a new objective function for  learning the neural control variates to alleviate the overfitting; and demonstrate its benefits in various  applications.  


\section{Constrained Neural Control Variates}
In this section, we propose constrained neural control variates (CNCV) for alleviating overfitting. Now we take a closer look at why the original objective function of NCV tends to bring a poor control variates if one optimizes the Eq.\eqref{eq:empvar} in the scenario that only a small number of samples from $p(\thetaB)$ are available. 

Firstly we note that the objective functions in Eq.~\eqref{eq:var}  and  Eq.~\eqref{eq:empvar} are not the same although $\mu_0$ is a constant, because the variance must be non-negative. For example, if we have a small number of samples, the learned neural network for $g$ could overfit the data so that the objective function in Eq.~\eqref{eq:empvar} could hit the global minimum $0$ due to the powerful capacity in approximation of the neural networks. Therefore, we have to optimize Eq.~\eqref{eq:empvar} with a constraint such that $\frac{1}{n}\sum_{i=1}^n [f(\thetaB_i) + g(\thetaB_i; \wB)]^2\geq\mu_0^2$. With this constraint, the solution would be $g(\thetaB_i;\wB) = - f(\thetaB_i)+\mu_0$ when using a small number of samples.
Without this constraint, we can easily observe that with a small $n$  and a large-capacity neural network for representing $\Phi(\thetaB; \wB)$, optimizing  Eq.~\eqref{eq:empvar} can easily result in ``point-wise'' fitting, $g(\thetaB_i;\wB) = - f(\thetaB_i)$, for each sample $\thetaB_i$, thus achieving  the minimal value of the objective. So it violates the constraint that the population mean of $g(\thetaB;\wB)$ is zero.  Therefore, directly minimizing Eq.~\eqref{eq:empvar} can cause severe overfitting. We thus propose two strategies for dealing with this issue. 

\begin{enumerate}
    \item {\bf Centering control variates}. Based on our analysis on optimizing Eq.~\eqref{eq:empvar}, it introduces bias for the true $g(\thetaB; \wB)$. To compensate this bias, we center the function $g(\thetaB; \wB)$ and set $g(\thetaB; \wB)= \widetilde{g}(\thetaB; \wB) - \mu$ where $\mu$ should be close to $\mu_0$. The parameter $\mu$ could also be learned during the training. Now if we substitute $g$ in Eq.~\eqref{eq:empvar}, the optimal function would be $\widetilde{g}(\thetaB_i;\wB) = - f(\thetaB_i)+\mu$ which would assure the required constraint $\frac{1}{n}\sum_{i=1}^n [f(\thetaB_i) + \widetilde{g}(\thetaB_i; \wB)]^2\geq\mu_0^2$. Note that in the following we assume $g$ is a centered function, and so denote $\widetilde{g}$ by $g$ for simplicity.
    \item {\bf Regularization}. We prefer a minimized variance of the function $g$. Thus the other strategy is to control the variance of the function $g$, $\Ebb[g^2]$, to regularize the complexity of the neural networks.    
\end{enumerate}
Combining the two strategies, the novel objective function can be formulated as the following,
\begin{equation}
\min_{\wB, {\mu}} \frac{1}{n}\sum_{i=1}^n \Big[ [f(\bm{\theta}_i) + g(\bm{\theta}_i; \bm{w}) - \mu]^2 + \lambda  g(\bm{\theta}_i ; \bm{w})^2 \Big], 
\label{eq:regncv}
\end{equation}
where $\lambda$ is the regularization parameter, and the  population variance $\Vbb[g]$ is estimated by its empirical samples as regularization term.

The random initialization of $\mu$ can slow down the training process and cause overfitting. To obtain a better performance, two optional initializing strategies could be used:
\begin{enumerate}
    \item Simply using $\frac{1}{n}\sum_{i=1}^{n}f(\bm{\theta}_i )$ as the initializing value of $\mu$;
    \item Pre-train the model with larger $\lambda$ till converged, then retain the value of $\mu$, randomly initializing other variables and re-train the model with smaller $\lambda$.
\end{enumerate}
When the number of samples $n$ is big enough, strategy 1 is recommended; when $n$ is small or $\Vbb(f(\bm{\theta}))$ is relatively large compared with $Ef(\bm{\theta})$, strategy 2 is recommended.


\section{Experiments}
\label{sec:app}
To evaluate our proposed method, we apply CNCV to a synthetic problem and two real scenarios, which are thermodynamic integration for Bayesian model evidence evaluation, and Bayesian neural networks. For comparison purposes, control functionals (CF)~\cite{oates2017control} and polynomial control variates~\cite{mira2013zero,oates2016controlled} are also applied to these problems.
The performance of the trained control variates are measured by the variance reduction ratio on the test data set, i.e.,
$$
\frac{\Vbb_{p(\bm{\theta})}[f(\bm{\theta}) + g(\bm{\theta})] }{\Vbb_{p(\bm{\theta})}(f(\bm{\theta}))]}.
$$
 We used fully connected neural networks to represent the trial function in all the experiments. We found that for the experiments presented in the following, a medium-sized network is empirically sufficient to achieve good performance. More details on network architectures are provided in Appendix.

\subsection{Synthetic Data}
To illustrate the advantage of NCV on dealing with high-dimensional problems over other methods, we consider to approximate the expectation of $f(\thetaB) = \sin(\pi /D \sum_{i=1}^D \theta_i )$ where $\thetaB \in \Rbb^D $ which is a mixture of Gaussians, i.e., $p(\thetaB) = 0.5\Ncal(-1, \IB) + 0.5 \Ncal(1,\IB)$.

\begin{figure*}[t!]
\centering
\vspace{-0.1cm}
\begin{tabular}{cc}
\includegraphics[width=0.5\columnwidth]{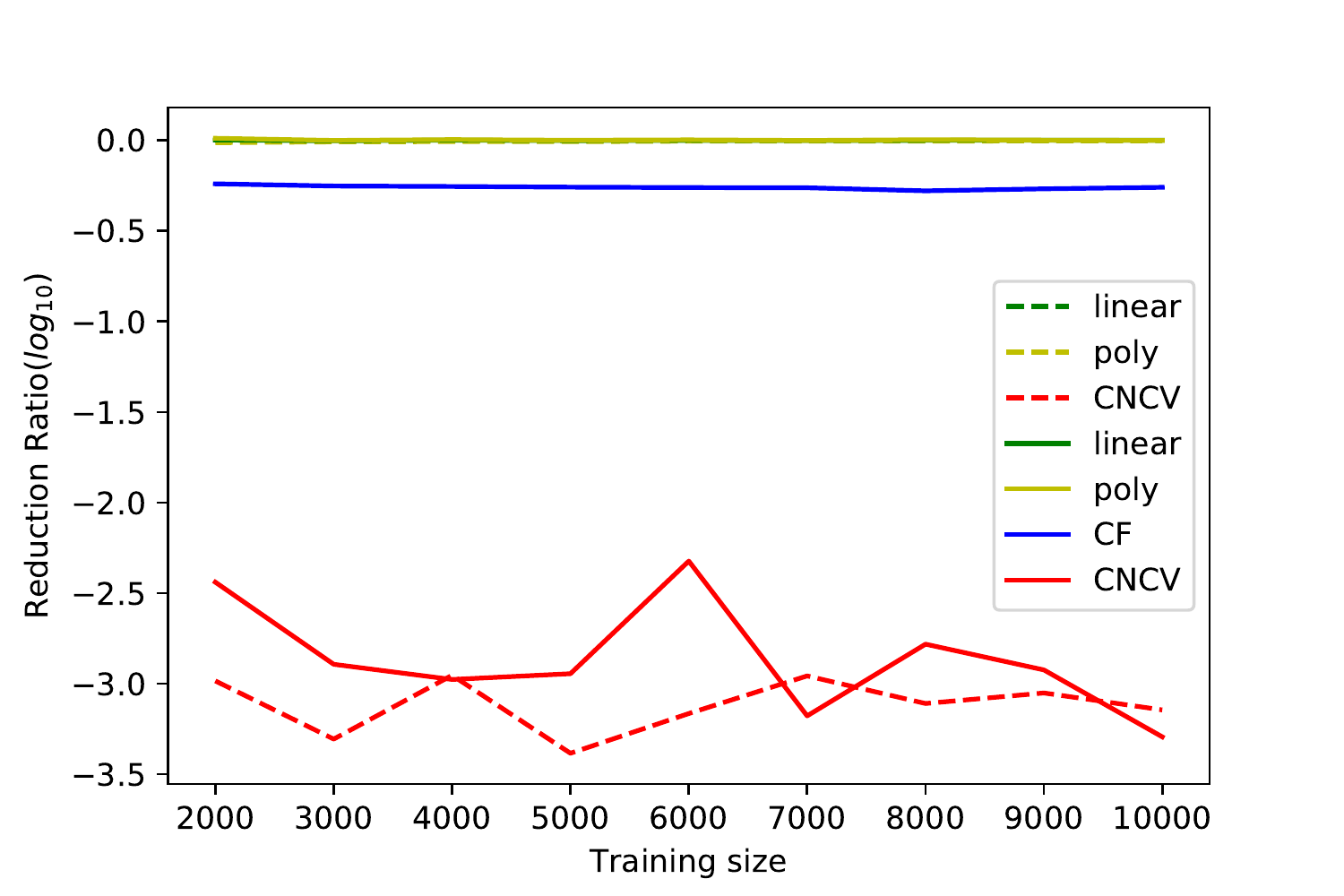}
\includegraphics[width=0.5\columnwidth]{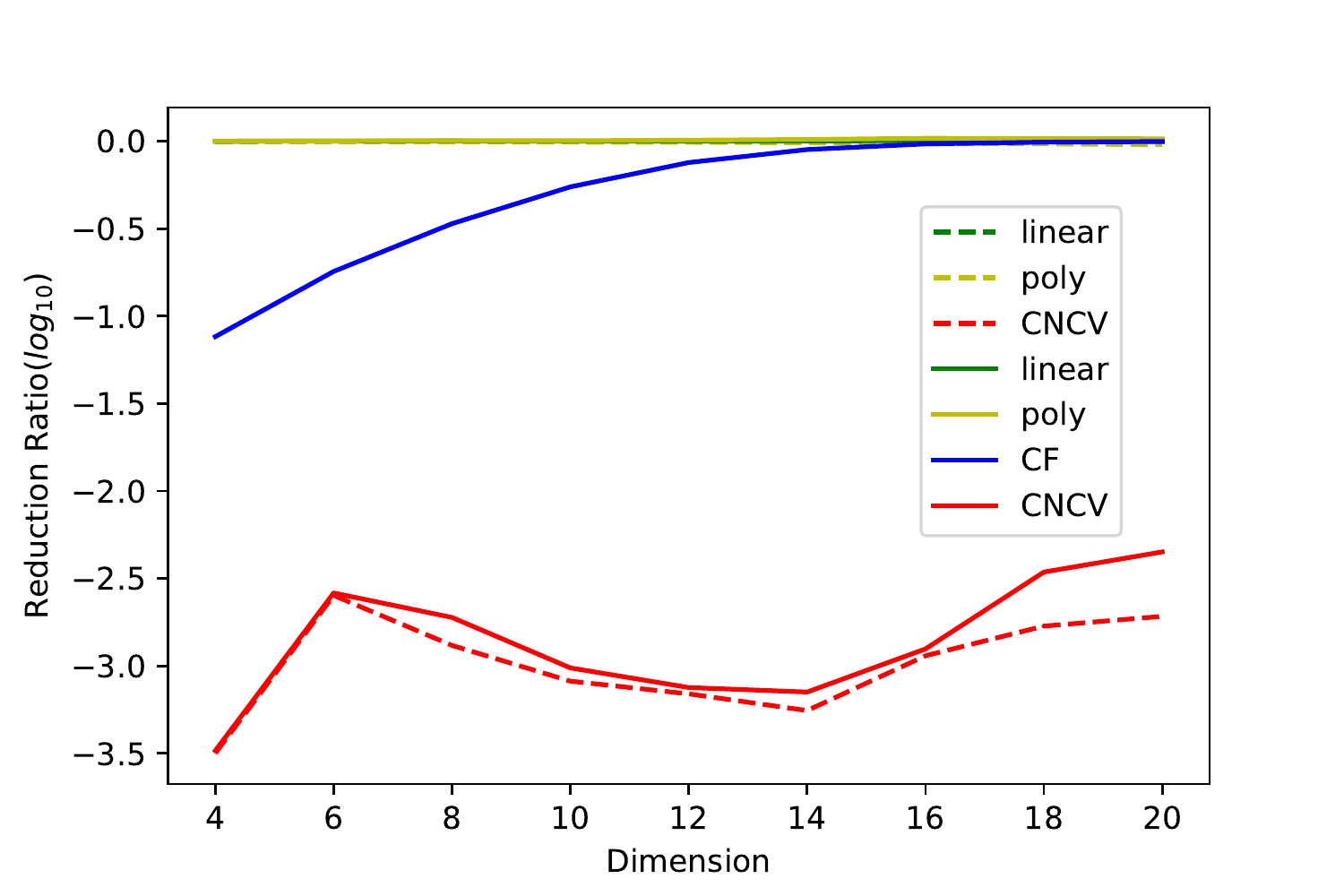}
\end{tabular}
\vspace{-0.2cm}
\caption{\small Synthetic data. (\textbf{Left}) Variance reduction ratio \emph{v.s.}  number of training  samples with $D=10$; (\textbf{Right}) Variance reduction ratio \emph{v.s.} dimension with training sample size $n=5000$.}
\label{fig:synthetic}
\end{figure*}

Figure~\ref{fig:synthetic} shows the variance reduction ratio on test data ($N=500$) with respect to varying the number of  training samples and the dimensions. In both cases, we can observe that CNCV outperforms linear, quadratic control variates and control functional. Particularly, when increasing the dimensions of $\thetaB$, CNCV can still achieve much lower variance reduction ratio compared with control functional. 

\begin{figure*}[ht!]
\centering
\includegraphics[width=0.8\textwidth]{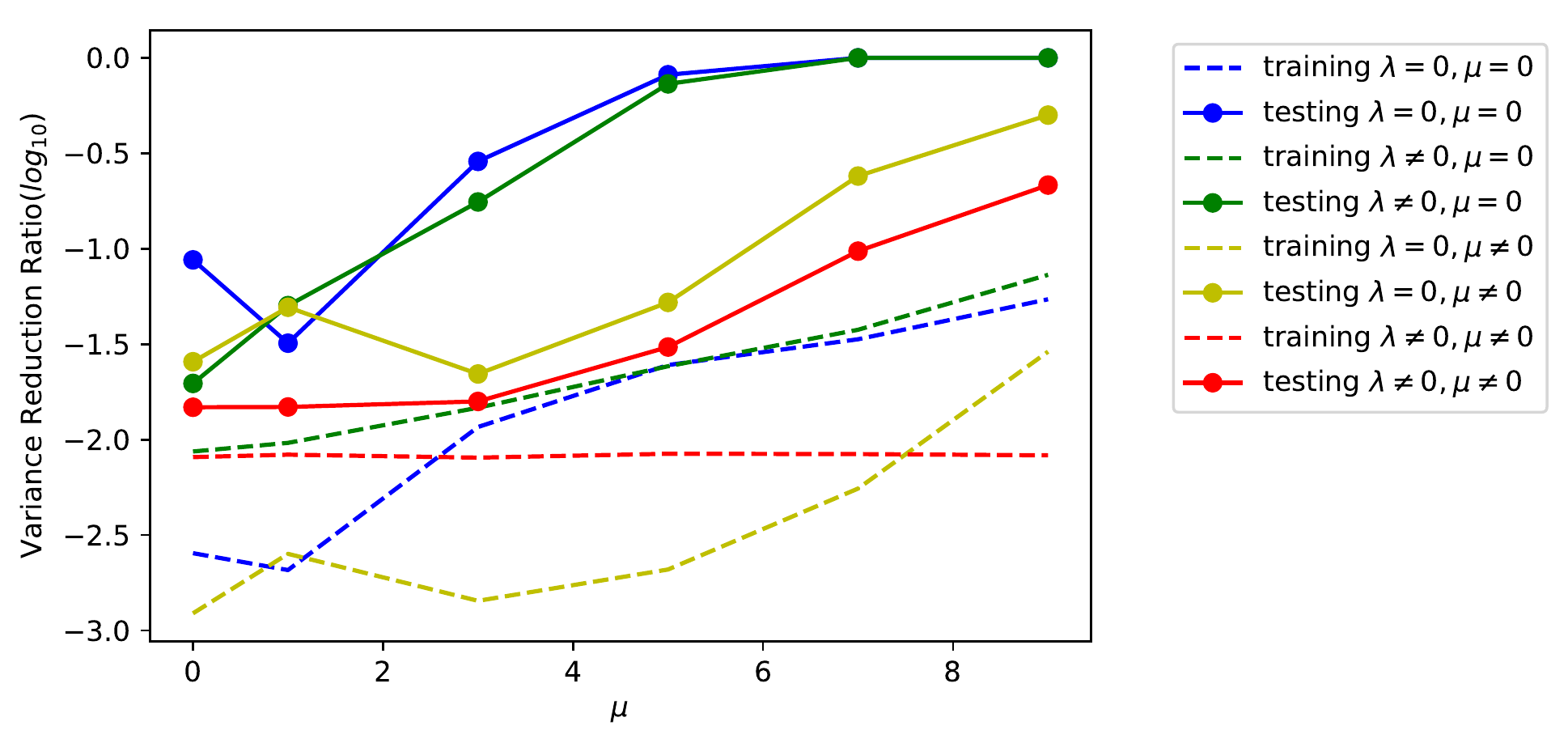}
\vspace{-0.45cm}
\caption{\small Variance reduction ratio of four types of NCV versus the oracle mean $\mu_0$. The $\mu$ in the control variates was initialized to $0$. Dashed and solid lines plot the results on training and test data respectively.}
\label{fig:toy_exp1}
\end{figure*}

\begin{figure}[ht!]
\centering
\vskip -0.1in
\includegraphics[width=0.7\columnwidth]{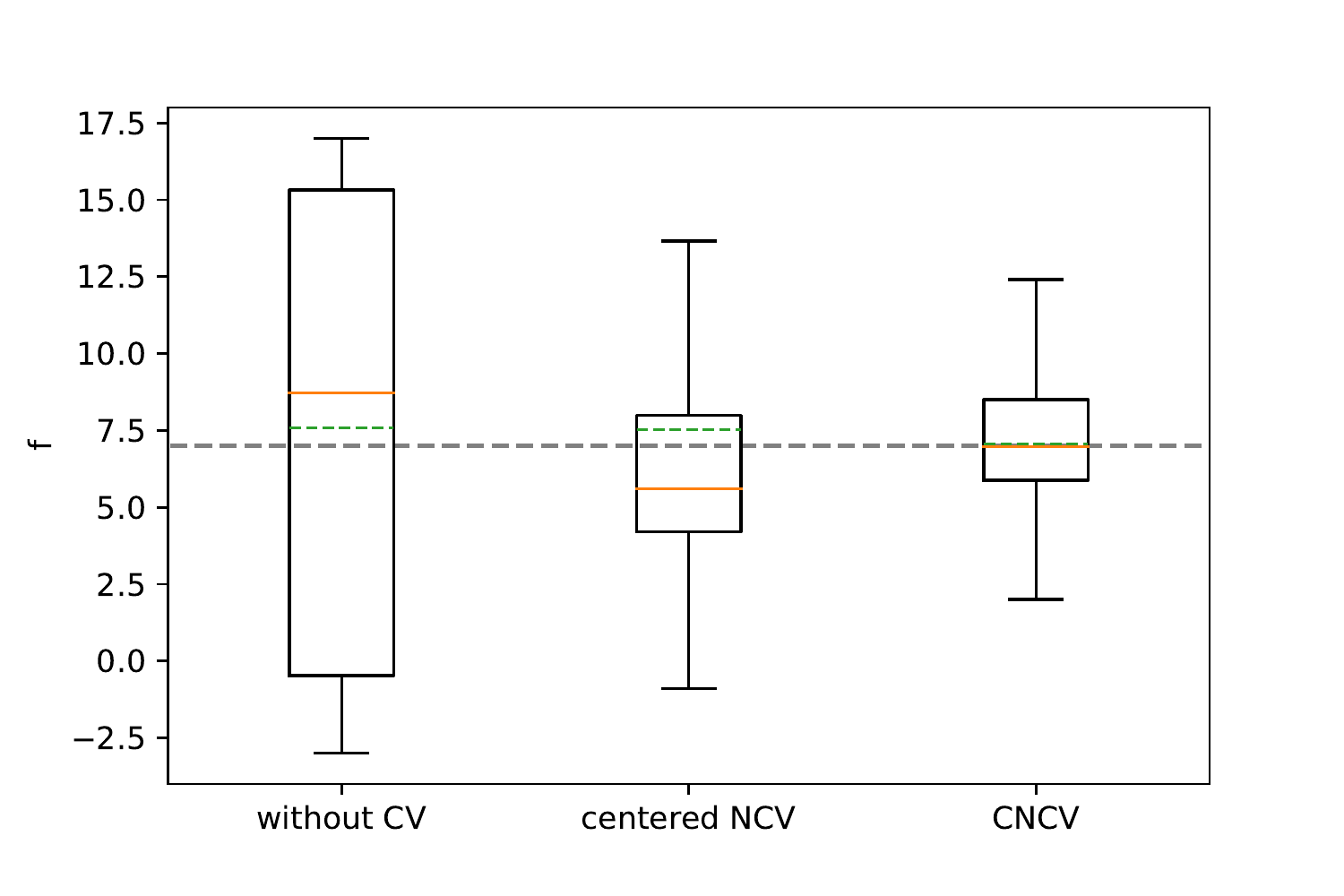}
\vspace{-0.45cm}
\caption{\small Boxplot of the samples for the function $f$ on test data. The orange solid line represents the median, the green dashed represents the sample mean, and the grey dashed line represents the oracle mean $\mu_0 = 7$.}
\label{fig:toy_exp2}
\end{figure}

Furthermore, we evaluated the two constraints made on the control variates in the Section 4. To highlight the comparison, we consider the modified function $f(\thetaB) = 10 \sin(\pi /D \sum_{i=1}^D \thetaB_i ) + \mu_0$ where $p(\thetaB) = 0.5\Ncal(-1, \IB) + 0.5 \Ncal(1,\IB)$, $\thetaB \in \mathbb{R}^{10}$. Here $\mu_0\in [0,9]$ represents the mean of $f(\theta)$, and $\sqrt{var(f(\theta))}\approx 7.5$. To evaluate our methods, we generated 1000 samples, where 500 samples were used for training and the rest were used for testing. Four neural control variates schemes with and without the constraints were applied to these samples. These schemes are: 1) not regularized, and not centered ($\lambda = 0, \mu=0$); 2) regularized, and not centered($\lambda \neq 0, \mu=0$); 3) not regularized, and centered($\lambda =0, \mu \neq 0$); 4) regularized, and centered($\lambda \neq 0, \mu \neq 0$).

Figure~\ref{fig:toy_exp1} reports the variance reduction ratio values for training and test data when varying $\mu_0$. It can be shown that NCV without constraints can easily be over-fitted with the training data. As $\mu_0$ increases, NCV with $\lambda = 0, \mu=0$ and NCV with $\lambda \neq 0, \mu=0$ were not able to reduce the variance for the test data. This shows that when $\mu_0$ is too large compared to the standard deviation, the control variates without constraints tends to fit $-f(\thetaB)$ rather than $-f(\thetaB) + \mu_0$ on the training data, which results in over-fitting. 

Figure~\ref{fig:toy_exp2} suggests that CNCV ($\lambda \neq 0, \mu \neq 0$) outperforms all the other methods. The NCV schemes with centered control variates ($\mu\neq 0$) were always better than the ones without centered control variates ($\mu = 0$). We can also see that the regularized control variates ($\lambda\neq 0$) can improve the performance. 

The $\mu$ was initialized to 0 in all experiments shown in Figure \ref{fig:toy_exp1}. To better understand the effect of the constraints on NCV, we reported the distribution of the samples $f(\thetaB)$ from test sets in Figure \ref{fig:toy_exp2}. It shows that although NCV, which was not regularized but centered ($\lambda =0, \mu \neq 0$), reduced the variance, the method does introduced bias so that the sample mean was away from the true mean $\mu_0$. The CNCV reduced the variance without introducing bias. 

In the following, we apply our proposed CNCV to two difficult problems with small number of samples. In these two cases, original NCV approach tends to severely overfit the training samples, leading to extremely poor generalization performance. Thus, we will not report the results of NCV.

\subsection{Thermodynamic Integral for Bayesian Model Evidence Evaluation}
In Bayesian analysis, data $\bm{y}$ is assumed to have been generated under a collection of putative models,  $\{\Mcal_i \}$. To compare these candidate models, the  Bayesian model evidence is constructed as 
$
p(\bm{y}|\Mcal_i) = \int p(\bm{y} | \bm{\theta}, \Mcal_i) p(\bm{\theta} |\Mcal_i) d\bm{\theta}
$ 
where $\bm{\theta}$ are the parameters associated with model $\Mcal_i$. Unfortunately, for most of the models of interest, this integral is unavailable in closed form.  Thus many techniques were proposed to approximate the model evidence. Thermodynamic integration (TI)~\cite{frenkel2001understanding} is among the most promising approach to estimate the evidence. This approach is derived from the standard thermodynamic identity, 
\begin{equation}
    \log p(\bm{y}) = \int_{0}^{1} \mathbb{E}_{p(\bm{\theta} | \bm{y}, t)} [\log p(\bm{y} | \bm{\theta} )]dt,
\end{equation}
where $p(\bm{\theta}|\bm{y}, t) \propto p(\bm{y}|\bm{\theta})^t p(\bm{\theta})$ ($t \in [0, 1]$) is called power posterior. Note that we have dropped the model indicator $\Mcal_i$ for simplicity. Here $t$ is known as an inverse temperature parameter. In many cases, the posterior expectation $\mathbb{E}_{p(\bm{\theta} | \bm{y}, t)} \log (p(\bm{y} |\bm{\theta}))$ can not be analytically computed, thus the Monte Carlo integration is applied. However, Monte Carlo integration often suffer large variance when sample size is not large enough. 

In~\cite{oates2016controlled}, the zero-variance control variates~\eqref{eq:laplacecv} were used to reduce the variance for TI, so that the posterior expectation is approximated by
\begin{equation}
\frac{1}{N}\sum_{i = 1}^N \log p(\bm{y} |\bm{\theta}_i^t) + \Delta Q_t(\bm{\theta}_i^t) + \nabla Q_t(\bm{\theta}_i^t) \cdot \nabla \log (p(\bm{\theta}_i^t |\bm{y}, t)
\end{equation}
where $\{\bm{\theta}_i^t \}_{i=1}^N$ are drawn from the posterior $p(\bm{\theta}|\bm{y}, t)$.  In \cite{oates2016controlled}, the trial function $Q_t(\bm{\theta})$ was assumed as a linear or quadratic function, which corresponds to a constant or linear function for $\Phi(\bm{\theta})$ in Stein control variates \footnote{We will call the trial function $Q(\bm{\theta})$ as the constant or linear type trial functions $\Phi(\bm{\theta})$ in the following.}. These methods achieved excellent performance for simple models~\cite{oates2016controlled}.  
However, they are struggling in some scenarios, for example, a negative example which is Goodwin Oscillator given in ~\cite{oates2016controlled}. Note that Goodwin Oscillator is a nonlinear dynamical system,
\begin{equation}
\frac{d \bm{x}}{ds} = f(\bm{x}, s; \bm{\theta}), \quad \bm{x}(0) = \bm{x}_{0}, \label{ode: TI}
\end{equation}
where the form of $f(\cdot)$ is provided in Appendix. 
Assuming within only a subset of time points $\{s_i \}_{i=0}^{N}$,  the solution of \eqref{ode: TI}, i.e. $\bm{x}(s_i, \bm{\theta})$, is observed under Gaussian noise $\varepsilonB(s) \sim \mathcal{N}(0, \sigma^2 \bm{I})$, where $\sigma^2$ denotes the variance of the noise. That means the observation $\bm{y}(s_i) = \bm{x}(s_i) + \bm{\varepsilon}(s_i)$.  Then we have the likelihood 
\begin{equation}
    p(\bm{y}|\bm{\theta}, \bm{x}_0, \sigma)  = \prod_{i=1}^N\mathcal{N}(\bm{y}(s_i)|\bm{x}(s_i; \bm{\theta}; \bm{x}_0), \sigma^2 \bm{I}).
\end{equation}
The expectation of log likelihood under the power posterior, i.e.,
$
\mathbb{E}_{p
(\bm{\theta}|\bm{y}, t)}\log p(\bm{y} | \thetaB),
$ 
needs to be evaluated. In \cite{oates2016controlled}, the authors demonstrated the failure of polynomial-type of control function since the log-likelihood surface is highly multi-modal and there is much weaker canonical correlation between the scores and the log posterior. 

In practice, sampling from Goodwin Oscillator is difficult and computationally expensive since simulating the underlying ordinary differential equation is extremely time-consuming. This directly leads to the situation that the available training samples for control variates are not sufficient. We show in the following that the proposed CNCV can be employed to deal with this issue.  
To illustrate the benefits of CNCV,  we compared it to other methods with various sizes of training samples and temperatures. For comparison purposes we evaluated the variance reduction ratios on both training and test sets ($500$ samples for test). The experiment settings are the same as those in \cite{oates2016controlled}.

\begin{figure*}[t!]
\centering
\vskip -0.1in
\begin{tabular}{cc}
\includegraphics[width=0.5\columnwidth]{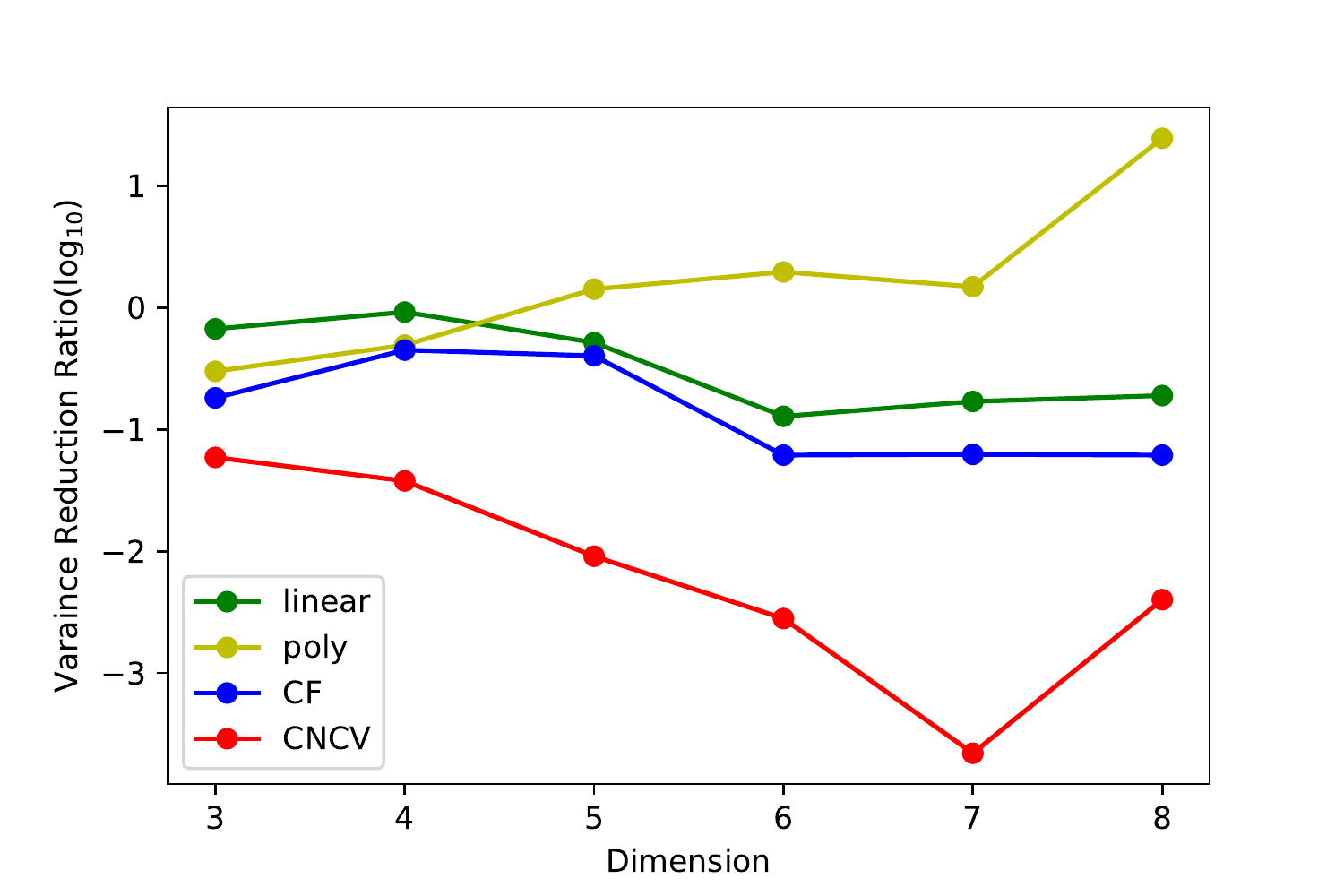} &
\includegraphics[width=0.5\columnwidth]{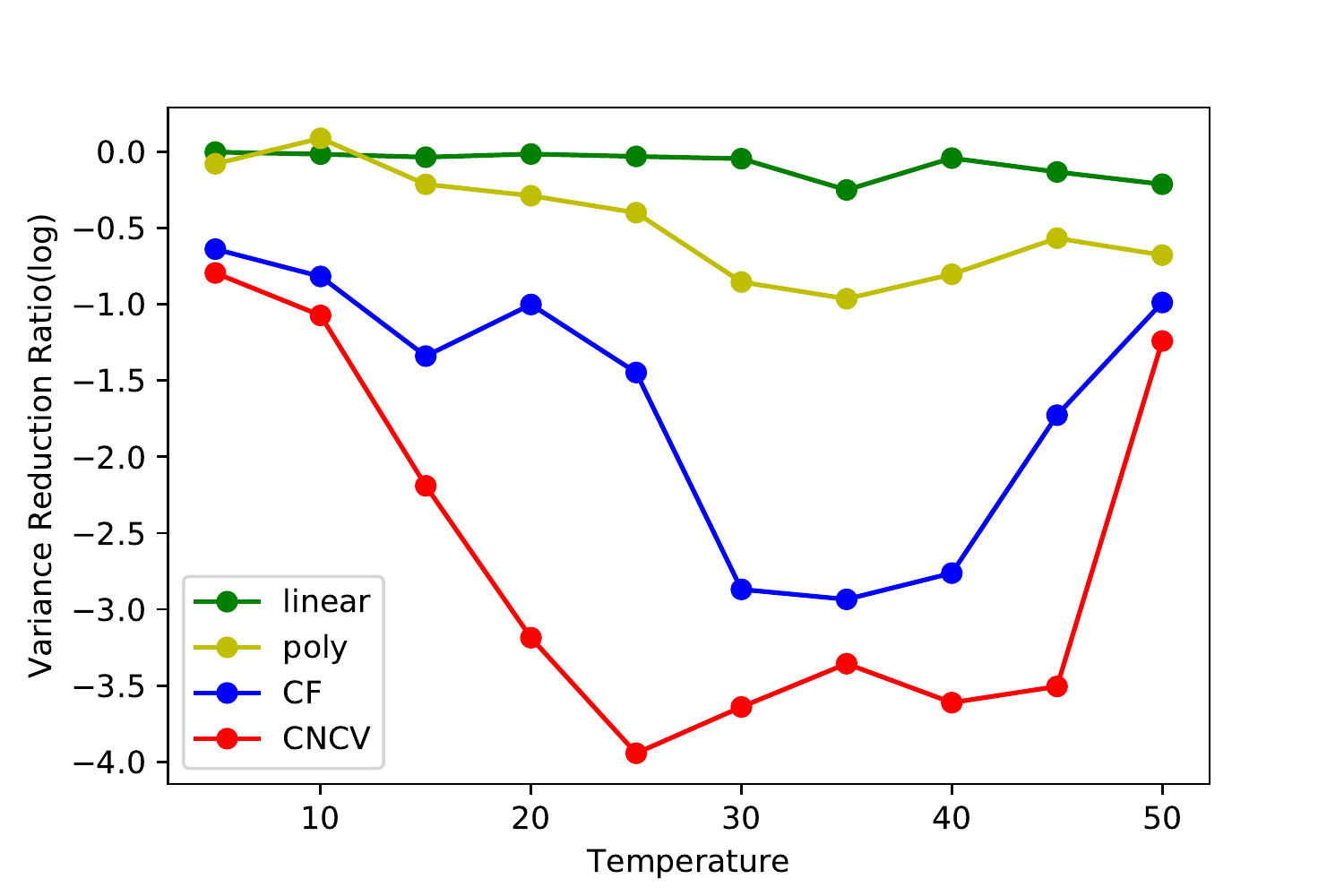} \\
(a)&(b)
\end{tabular}
\vspace{-0.3cm}
\caption{Variance reduction ratio on test set of four different types of control variates (linear, quadratic, CF and CNCV). 3000 samples were used for training and the other 3000 samples were used for testing. (a) The average variance reduction ratio on test data versus the problem dimension; (b) The average variance reduction ratio on the test data for different temperatures.}
\label{fig:ti}
\end{figure*}

Figure~\ref{fig:ti} shows the experimental results when applying different types of control variates. It can be easily observed that the linear and quadratic methods could hardly reduce the variance of the Goodwin Oscillator model on testing set, while CNCV obtained the lowest variance reduction ratio comparing to all other methods. Control functional can significantly reduce the variance when dimension is low, but CNCV still can get the lowest variance reduction ratio, inspite of the problem dimensions or temperatures.

\subsection{Uncertainty Quantification in Bayesian Neural Network}
Standard neural network training via optimization is equivalent to maximum likelihood estimation (MLE for short). Given the training samples, $\{(\bm{x}_i, \bm{y}_i)\}_{i=1}^N$, and denote $\XB = \{ \xB_i\}_{i=1}^N$ and $\YB = \{ \yB_i\}_{i=1}^N$, the weight parameter $\bm{\theta}$ of the neural networks is estimated by
\begin{equation}
\hat{\bm{\theta}} = \argmax_{\bm{\theta} } \sum_{i=1}^N \log p (\bm{y}_i |\bm{x}_i; \thetaB)
\end{equation}
However, the solution of MLE lacks theoretical justification from the probabilistic perspective to deal with the parameter uncertainty as well as structure uncertainty. Moreover, standard neural networks are  often susceptible to producing  over-confident predictions. Bayesian natural network~\cite{neal2012bayesian} was introduced for implementing uncertainty quantification. 
Firstly, one provides a prior distribution over the weights, 
$p_0(\bm{\theta}) = \mathcal{N}(\bm{0}, \sigma_0^2 \bm{I}) $, 
where $\sigma_0^2$ is the variance magnitude. Assuming the likelihood function has the form, $p(\yB_i | \xB_i; \thetaB) = \Ncal \left( \yB_i | NN(\xB_i; \thetaB), \sigma^2 \IB \right)$, then the posterior of the weights
$\thetaB$, 
$$
p(\bm{\theta}| \bm{X}, \bm{Y}) \propto \prod_{i=1}^N p(\bm{y}_i |\bm{x}_i, \bm{\theta})p_0( \bm{\theta}).
$$ 
The uncertainty of the model, typically formulated as the expectation of a specific statistics $f(\bm{\theta},\bm{x}, \bm{y})$ could be computed based on the posterior distribution of the weights, 
\begin{equation}
\mu_{f}  = \Ebb[f(\thetaB;\xB, \yB)] = \int f(\thetaB;\xB, \yB) p(\bm{\theta}| \bm{X}, \bm{Y}) d \thetaB
\end{equation}

Due to the analytic intractability of the integral, the expectation of $f(\bm{\theta}, \bm{x}, \bm{y})$ is estimated using Monte Carlo integration
\begin{equation}\label{BNNest}
\mu_{f}  \approx \frac{1}{M}\sum_{i=1}^M f(\bm{\theta}_i, \bm{x}, \bm{y})
\end{equation}
where $\{ \bm{\theta}_i \}_{i=1}^M$ is drawn from the posterior $p(\bm{\theta} | \bm{X}, \bm{Y})$. 

However, the large number of parameters and complex structure of networks make the sampling from the posterior extremely hard. Typically, only a small number of samples could be obtained.  Consequently, small sample size and the complex structure of the posterior distribution will lead to a high variance of the estimator \eqref{BNNest}. Thus, we consider reducing the variance of Monte Carlo estimator by NCVA. 

\paragraph{Uncertainty quantification on predictions with out-of-distribution inputs.} The neural networks learned with the MLE principal could achieve a high-accuracy performance, when the training data and test data come from the same data distribution. But when an out-of-bag (OOB) sample, i.e., a sample whose label is not included in the training set, is fed into the models, the MLE model is very likely to identify the OOB sample as a certain in-bag class with very high confidence, i.e. the prediction score is close to 1. We hope to construct a robust classifier which won't misclassify the OOB samples with very high confidence. The Bayesian neural network is considered to be effective to handle this situation. However, evaluating the expected prediction score under the posterior of $\wB$ still suffers large variance issues. Hence we considered to reduce the variance of BNN prediction score and handle the over-confident issues of OOB samples.

We implemented a simple image classification task to evaluate the effectiveness of CNCV. We selected all the images with label "6" and "9" from the MNIST dataset and constructed a convolutional neural network for Bayesian classifier with the output $\{f(\bm{x}, \bm{\theta}_i)\}_{i=1}^M$ as the probability of class assignment, where $\bm{\theta}_i \sim p(\bm{\theta} | \bm{X}, \bm{Y})$ on the two categorizes and select the images with label "8" as the out-of-distribution samples $\xB_{\text{out}}$ for test.  
We constructed the control variates to reduce the variance of the estimator $\hat{P}(y = \text{``6''} |\xB_{\text{out}} )$ using NCV,
\begin{align}\label{BNNest: oob}
\hat{P}(y = \text{``6''}|\xB_{\text{out}}) &= \frac{1}{M}\sum_{i=1}^M \hat{P}(y = \text{``6''}|\xB_{\text{out}}, \thetaB_i) + g(\bm{\theta}_i, \xB_{\text{out}})
\end{align}

The  parameters $\{ \bm{\theta}_i \}$  are sampled based on the training set, and hyperparameters are tuned using the validation set. Both the training and validation data are composed of the images with labels ''6'' or ''9''. We evaluated the control variates methods on the test set, consisting of the images with the label ``8''. We evaluate the control variates by computing the following variance ratio
\begin{equation}
\frac{1}{N}\sum_{i=1}^N \frac{\Vbb_{p(\bm{\theta}|\bm{X}, \bm{Y})}[p(y_i = \text{"6"}| \xB_i) + g(\xB_i, \bm{\theta})]}{\Vbb_{p(\bm{\theta}| \bm{X}, \bm{Y})}[p(y_i = \text{"6"}| \xB_i)]}
\end{equation}

Due to the high dimensionality of this problem, quadratic control variates and control functional failed to obtain satisfying variance reduction, and we thus do not report their results.

\begin{figure*}[ht!]
\centering
\vskip -0.1in
\begin{tabular}{cc}
\includegraphics[width=0.5\columnwidth]{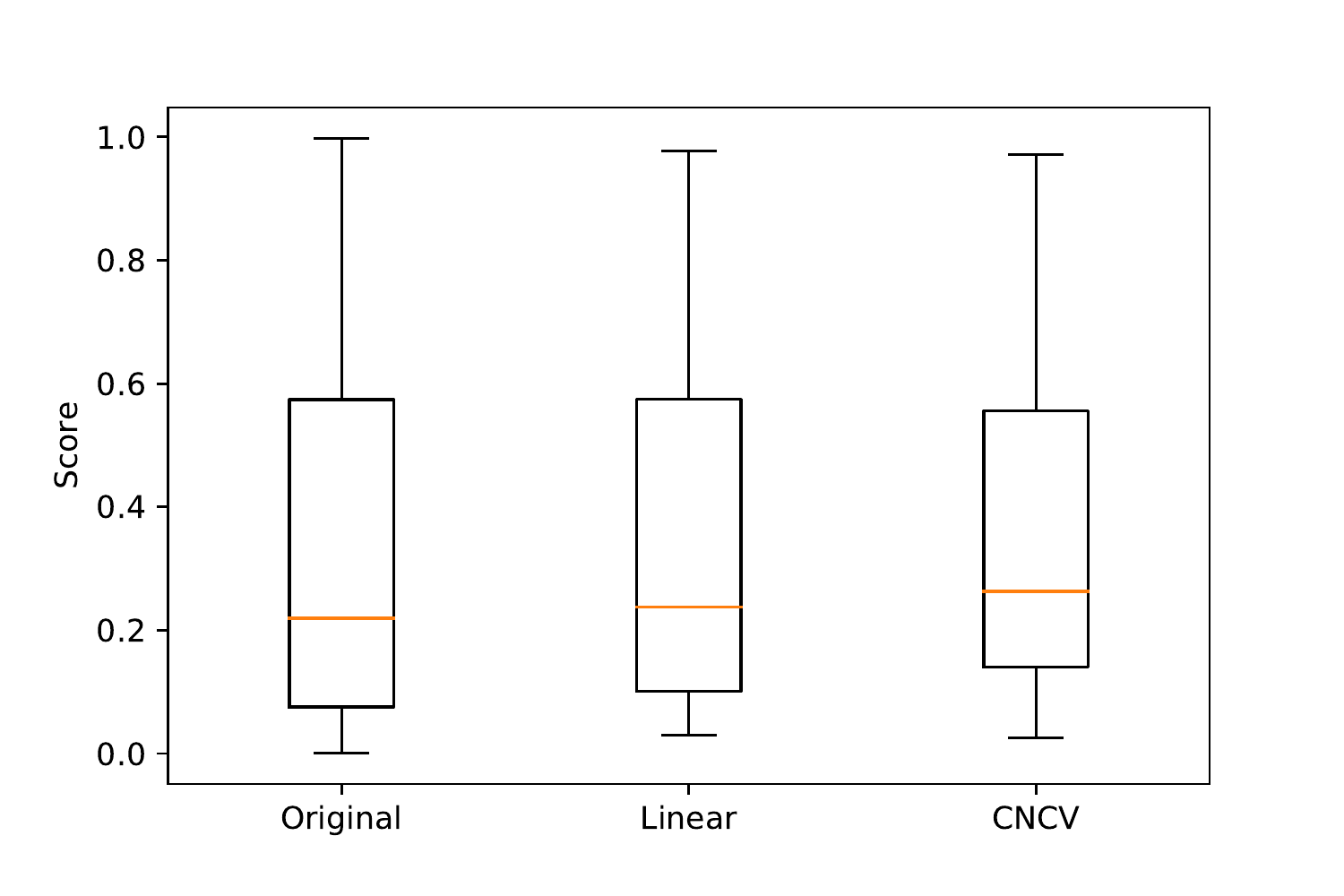} & 
\includegraphics[width=0.5\columnwidth]{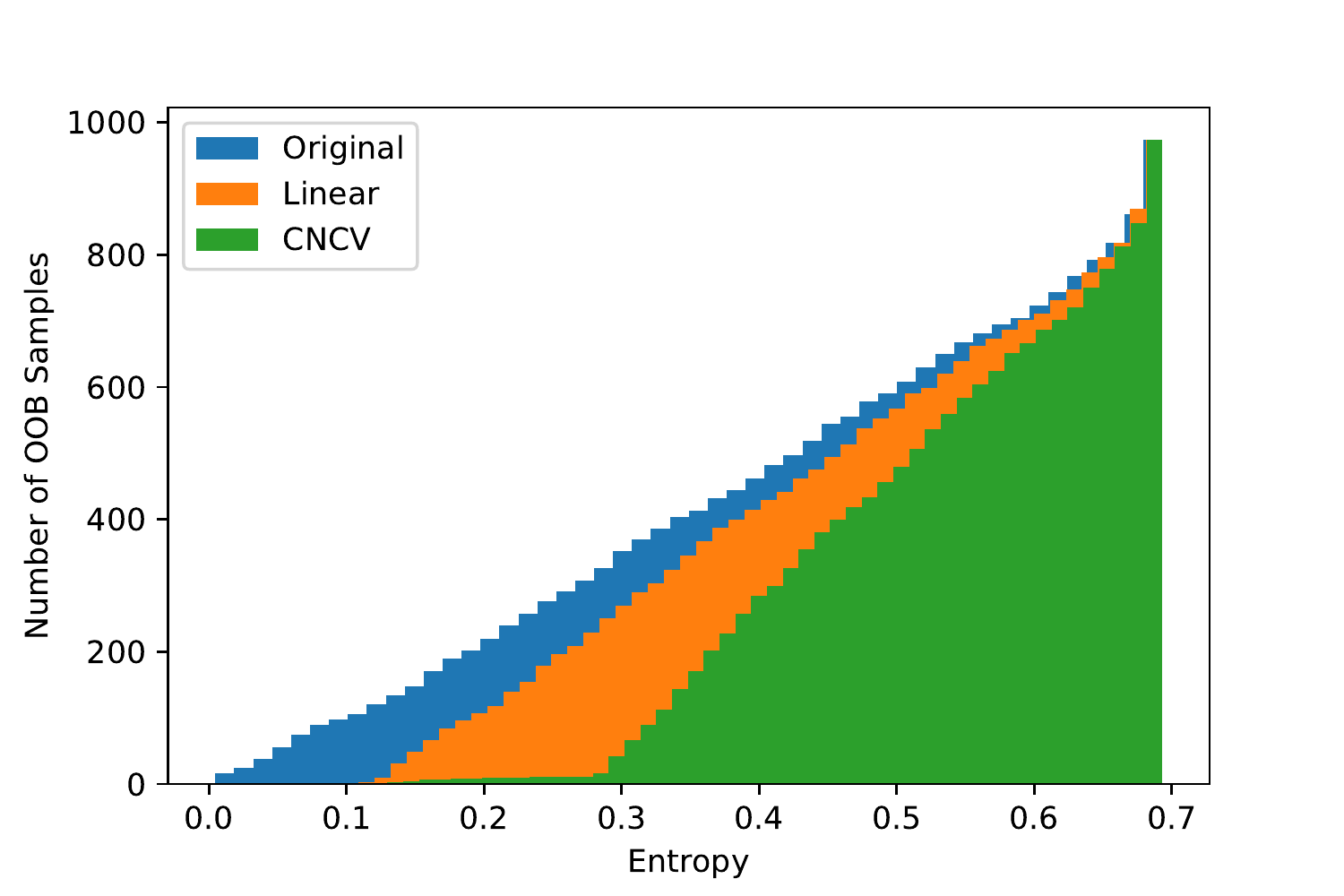} \\
(a) & (b)
\end{tabular}
\vspace{-0.2cm}
\caption{Performance of CNCV on BNN with OOB samples (a) boxplot of the prediction score of OOB samples defined in Equation \eqref{BNNest: oob}. The prediction scores were computed based on BNN ensemble classifier for those variance reduction methods. (b) The accumulated empirical distribution of the entropy computed by prediction scores using Equation \eqref{entropy}.}
\label{fig:bnn}
\end{figure*}

Figure~\ref{fig:bnn}(a) shows that the overall  distribution of BNN ensemble prediction does not change significantly, where CNCV produces slightly better results. This is expected since the classifier has not seen the OOB samples during training, which makes it impossible to yield a stable prediction probability. On the other hand, Figure~\ref{fig:bnn}(b) depicts the the entropy of these OOB samples computed using prediction score via BNN:
\begin{equation}\label{entropy}
    Entropy(x_{OOB}) =  - \hat{p} \log (\hat{p}) - (1-\hat{p})\log (1 - \hat{p}),
\end{equation}
where $\hat{p}$ is the BNN prediction score evaluated from Eq.~\eqref{BNNest: oob}. $Entropy(x_{OOB})$ is in the range $[0, \log 2]$. Samples with low entropy close to $0$ means they will be classified as $6$ or $9$ with very high confidence. It could be seen from Figure~\ref{fig:bnn}(b) that BNN prediction with control variates has less over-confident scores over OOB samples. That means that BNN prediction with CNCV yields the least over-confident scores compared with vanilla BNN and that with linear control variates.


\section{Conclusion}
We have proposed neural control variates for variance reduction. We have shown that the neural control variates could have the over-fitting problem when using a small number of samples. To alleviate this over-fitting problem, we proposed constrained neural control variates, where the control variates is centered and regularized. We demonstrated the effectiveness of the proposed methods on synthetic data and  two challenging  Monte Carlo integration tasks. However, the theoretical justification of the proposed method will be investigated in our future research. 


\appendix

\section{Formulas for Goodwin Oscillator}
The nonlinear dynamic system of the Goodwin Oscillator used in \cite{oates2016controlled} is given by:
\begin{equation}
\begin{split}
\frac{d x_1}{d s} &= \frac{a_1}{1+ a_2 x_g^{\rho}} - \alpha x_1 \\
\frac{d x_2}{d s} &= k_1x_1 - \alpha x_2 \\
    & \vdots \\
\frac{d x_g}{d s} &= k_{g-1} x_{g-1} - \alpha x_g.
\end{split}
\end{equation}

The solution $\bm{x}(s; \theta, \bm{x}_0)$ of this dynamical system depends on the uncertain parameters $\alpha, a_1, a_2, k_1,\dots, k_{g-1}$. Similar to the settings in \cite{oates2016controlled}, we assume $\bm{x}_0 = [0, \dots, 0]$ and $\sigma = 0.1$ are both known and take sampling times to be $s = 41,\dots,80$. Parameters were assigned independent $\Gamma(2, 1)$ prior distributions. We generated data using $a_1 = 1, a_2 = 3, k_1 = 2, k_2, \dots ,k_{g-1} = 1, \alpha = 0.5$. We generated the posterior samples of the weights using MCMC with parallel tempering. In each dimension cases ($g \in \{3, 4, \cdots, 8\}$) the Markorv Chain runs 100, 000 iterations to ensure converge, 6000 samples randomly drawn from the last 50, 000 iterations were used in the final experiments. \\

The trial function $\phi$ used in Goodwin Oscillator is a two layers fully connected neural network, where each layer has 40 neurons. The activation function is the Sigmoid function. 

\section{Uncertainty Quantification in Bayesian Neural Network: Out-of-Bag Sample Detection}
The basic model consists of two convolutional layers, two max-pooling layers and a fully connected layers, with kernel size $(5 \times 5 \times 2)$, $(2\times 3\times 3 \times 3)$, $(147 \times 2)$ respectively. The prior distribution of the weight was set to standard normal distribution $\mathcal{N}(0, 1)$. The samples of the weights were generated using preconditioned Stochastic Gradient Langevin Dynamic \cite{li2016preconditioned}. 1000 samples were generated to construct the Bayesian neural network prediction. The trial function $\phi(\bm{\theta}, x): \Theta \times \mathbb{X} \longrightarrow \mathbb{R}$ was defined as:
\begin{equation}
    \phi(\bm{\theta}, x) = \alpha^T h(W_0\bm{\theta} + \psi(x))
\end{equation}
where $\psi(x)$ consists of two convolutional layers with kernel size $(5 \times 5 \times 2)$, $(2\times 3\times 3 \times 3)$, two max-pooling layers and relu activation. $W_0 \in \mathbb{R}^{147 \times 407}$, $h$ is the sigmoid function. and $\alpha \in \mathbb{R}^{147}$. Thus the neural control varaites of the BNN prediction is:
\begin{equation}
   g(\bm{\theta}, x) =   \nabla_{\bm{\theta}} \cdot \phi(\bm{\theta}, x) +  \phi(\bm{\theta}, x) \nabla_{\bm{\theta}} \cdot \log p(\bm{\theta} | \bm{X})
\end{equation}
where $\nabla \cdot f = \sum_{i} \frac{\partial f}{\partial x_i}$.

\end{document}